\pdfoutput=1
\documentclass[11pt]{article}
\usepackage[preprint]{acl}
\usepackage{tabularx}
\usepackage{times}
\usepackage{threeparttable}
\usepackage{latexsym}
\usepackage{tablefootnote}
\usepackage[toc,page]{appendix}
\usepackage[T1]{fontenc}

\usepackage[utf8]{inputenc}
\usepackage[russian,english]{babel}
\usepackage{microtype}

\usepackage{inconsolata}

\usepackage{graphicx}
\usepackage{svg}
\usepackage{amsmath}

\title{TEII: Think, Explain, Interact and Iterate with Large Language Models to Solve Cross-lingual Emotion Detection}

\author{Long Cheng\thanks{These authors contributed equally to this work.}, Qihao Shao\footnotemark[1], Christine Zhao, Sheng Bi, Gina-Anne Levow\thanks{Corresponding author}\\
Department of Linguistics \\
  University of Washington \\
  \texttt{\{lcheng97, qihaos, czhao028, shengbi, levow\}@uw.edu} \\}

\begin{document}
\maketitle
\begin{abstract}
Cross-lingual emotion detection allows us to analyze global trends, public opinion, and social phenomena at scale. We participated in the Explainability of Cross-lingual Emotion Detection (EXALT) shared task, achieving an F1-score of 0.6046 on the evaluation set for the emotion detection sub-task. Our system outperformed the baseline by more than 0.16 F1-score absolute, and ranked second amongst competing systems. We conducted experiments using fine-tuning, zero-shot learning, and few-shot learning for Large Language Model (LLM)-based models as well as embedding-based BiLSTM and KNN for non-LLM-based techniques. Additionally, we introduced two novel methods: the Multi-Iteration Agentic Workflow and the Multi-Binary-Classifier Agentic Workflow. We found that LLM-based approaches provided good performance on multilingual emotion detection. Furthermore, ensembles combining all our experimented models yielded higher F1-scores than any single approach alone. 

\end{abstract}

\section{Introduction}
In this study, we focused on tackling the cross-lingual emotion detection task for Tweets, which is a sub-task in EXALT@WASSA 2024 \citep{Maladry2024}. This task is interesting for its global application in understanding emotions across languages. It is also challenging due to linguistic diversity and cultural differences in emotional expression. To tackle the multilingual challenge, we conducted experiments using multilingual LLM-based models as well as classical machine learning models that used multilingual embeddings. An innovation developed within these experiments is the creation of an Agentic Workflow approach that leverages the strengths of multiple LLMs for the emotion detection task. All code will be released on GitHub\footnote{\href{https://github.com/cl-victor1/EXALT_2024_BCSZ}{https://github.com/cl-victor1/EXALT\_2024}}. 
%@inproceedings{zhou-wu-2018-nlp,
%yin-shang-2022-efficient
\section{Related Work}
Previously, research like that of \citet{hassan-etal-2022-cross} explored classification using classical models such as BERT and SVMs, trained with various linguistic features. More recently, \citet{thakkar-etal-2024-m2sa-multimodal} investigated sentiment recognition in tweets using both multimodal and multilingual approaches. 

ChatDev by \citet{qian2023communicative}, Gorilla by \citet{patil2023gorilla}, HuggingGPT by \citet{shen2023hugginggpt}, and the Reflexion framework by \citet{shinn2023reflexion} highlighted the potential of multi-LLM-agent collaboration, termed Agentic Workflow,  in solving complex tasks and its application to tool use, code generation, and similar activities.   A related approach is AutoGen \citep{wu2023autogen}, where natural language and computer code are integrated to tackle complex tasks that are challenging for a single prompt or one LLM. We believe this broad methodology can be applied to the emotion detection task in a multilingual setting. In this work, we introduce Agentic Workflow 
using a multi-agent approach to enhance the performance of individual LLMs in detecting emotions in tweets.

\begin{figure*}[h!]
\centering
  \includegraphics[width=\textwidth]{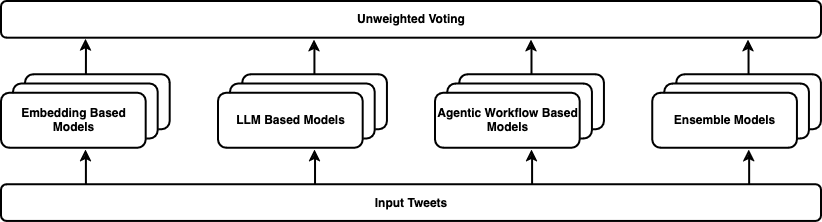}
  \caption{System architecture of the final ensemble model that combines both individual models and other ensemble models with fewer individual models.}
  \label{Architecture}
\end{figure*}

\section{System Description}
We explored three broad classes of models for the EXALT cross-lingual emotion recognition tasks. One approach trained several KNN models and a BiLSTM model with multilingual embeddings to encode EXALT's cross-linguistic Tweet data directly (detailed in Appendix \ref{A}). Another group of experiments employed LLM-based models weakly or not directly trained for EXALT, through prompting. For the third class of approaches, we developed two Agentic Workflow methods, using multiple agents: Multi-Iteration Agentic Workflow and Multi-Binary-Classifier Agentic Workflow. Finally, we applied ensemble methods to aggregate across approaches. The system architecture we used for the final evaluation is illustrated in Figure \ref{Architecture}.

\subsection{Prompt Based Classification with LLMs}
With the advancement of LLMs, formulating natural language processing problems into text completion problems via prompting has shown promising results. We explored fine-tuning OpenAI's GPT3.5, a zero-shot setting using OpenAI's GPT4\footnote{gpt-4-turbo-2024-04-09 and gpt-4o-2024-05-13} and Anthropic's Claude3\footnote{claude-3-opus-20240229}, and a few-shot setting using OpenAI's GPT4. Additionally, leveraging the Chain-of-Thought \citep{wei2023chainofthought} approach, we designed two more methods built upon the zero-shot model by explicitly asking for explanation from LLMs in their outputs. All detailed prompts that we used for different models can be found in Appendix \ref{C}.

\paragraph{Fine-tuning GPT3.5 (FineTuneGPT)}
We partitioned the training dataset into a training dataset and a validation dataset, comprising 4000 and 1000 instances respectively, and used the datasets to create a fine-tuned GPT3.5 model. For inference, the system prompt remained consistent with that utilized during the fine-tuning process.

\paragraph{Zero-Shot (ZeroShot)}
Under the zero-shot setting, we set the system prompt to align with our task goal and then directly asked LLMs to output a label among the six emotion labels given a certain tweet.
\paragraph{Zero-Shot with Explanation (ZSE) and Correction (ZSEC)}
Building upon the zero-shot setting, we asked LLMs to provide an explanation before assigning emotion labels during inference (ZSE). Taking the Chain-of-Thought idea one step further, we introduced a second LLM in the same inference process, and used the output label from the first LLM as part of the input to the second LLM (ZSEC). More specifically, we asked the first LLM to explain and output the emotion label, and then we asked the second LLM to check whether it agreed with the label output by the first LLM. If the second LLM agreed with the first LLM, it would output the same label; otherwise, it would output an alternative label. In either case, the second LLM would also provide explanation before outputting the emotion label. With this approach, we take the output from the second LLM as the final output.
\paragraph{Few-Shot (FewShot)}
We set the system prompt similar to that of the zero-shot setting and provided a few example tweets with their associated emotion labels before asking LLMs to output a label for a certain tweet. We employed both random sampling and embedding-based KNN (with $k=6$) to pick example tweets from the training dataset. 

\subsection{Agentic Workflow (AWF)}
Drawing inspiration again from the concept of Chain-of-Thought \citep{wei2023chainofthought} and the multi-agent conversation framework outlined in \citep{wu2023autogen}, we developed two Agentic Workflow methods to enhance the performance of LLMs. 
% The agentic workflow entails collaboration among two or more AI agents to address intricate tasks.

\paragraph{Multi-Iteration Agentic Workflow (MIAWF)}
This approach involves using one or more LLM agents to adjudicate between the outputs of prior models. First, we identify the two top-performing models (agent 1 and agent 2) based on their respective F1-scores on dev data. Following this selection, an additional LLM is introduced as Agent 3. We prompted agent 3 to assess the outputs of Agents 1 and 2 and select the optimal label. In this manner, the classification decision is reduced to binary from multi-class, and leverages the output of strong models.

\begin{figure}[h!]
\centering
  \includegraphics[scale=0.25]{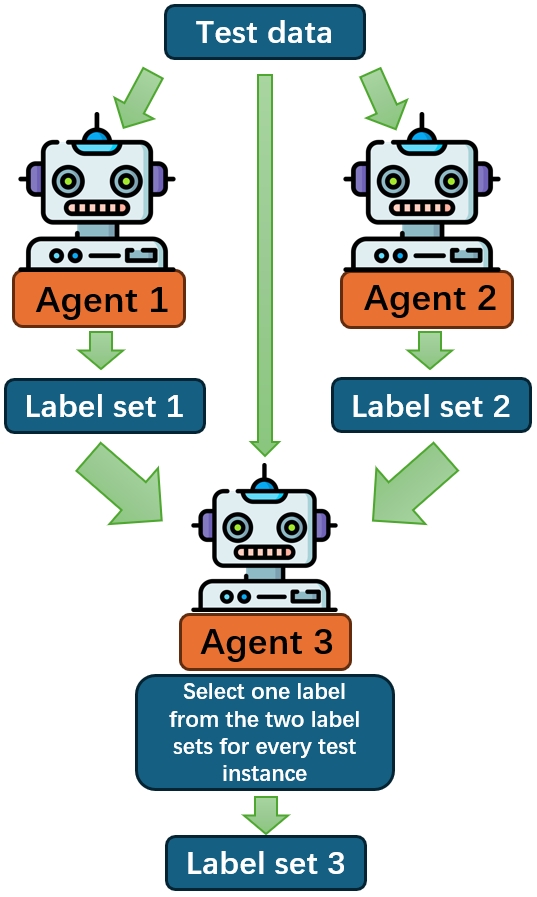}
  \caption{One iteration of Multi-Iteration Agentic Workflow. (This figure has been designed using images from Flaticon.com)}
  \label{AW_graph}
\end{figure}

After obtaining the results for agent 3, another LLM could be introduced as agent 4. The system prompt for agent 4 would remain the same as that for agent 3. However, this time, the source models for agent 4 are agent 3 and the better of agent 1 and agent 2. It is observed that agent 4 often outperforms agent 3 slightly on the dev set. In theory, this iteration could be repeated multiple times, but as the source models gradually become more similar to each other, the performance improvement may diminish. One iteration of MIAWF is presented in Figure \ref{AW_graph}.

\paragraph{Multi-Binary-Classifier Agentic Workflow (MBCAWF)}
This approach is inspired by both the idea of ensemble learning and the previous Agentic Workflow approach. First of all, we hypothesized that LLMs would have better performance on binary classification than on multi-class classification. Since there were six different emotions in Task 1, we made five binary classifiers, one for each of the emotions except for the emotion "Neutral". Secondly, we observed that LLMs had performed well in selecting the preferred output above. We thus extended the Agentic Workflow further and combined it with the binary classifiers.

Multi-Binary-Classifier Agentic workflow works as follows for inference on each instance:
\begin{enumerate}
\item Ask the LLM whether the tweet has each of the non-neutral emotions (Binary Classifiers).
\item If only one of the emotions is predicted positive, use that emotion as the predicted emotion.
\item If multiple emotions are predicted positive, ask the LLM to pick one among the positive ones as the predicted emotion.
\item If there is no emotion detected (all emotions predicted negative), tell the LLM that others think the tweet is of "Neutral" emotion and ask it to double check that classification. If so, output "Neutral" as the predicted emotion; otherwise, pick one among the other five emotions as the predicted emotion.
\end{enumerate}
With this approach, we expect that both precision and recall are going to be improved, especially for non-neutral emotions.

\subsection{Ensembles}
Due to noticeable variations in outputs from our base models, primarily LLMs, we hypothesized that consolidating predictions through an ensemble mechanism would yield beneficial results. Consequently, during the development phase, we evaluated each model architecture at various hyperparameters to choose the highest dev-set F1-score version to include in our ensembles. Comparative analyses of these methods are provided in Table \ref{ensemble_types} in Appendix \ref{B}. Since the unweighted voting ensemble demonstrated the best performance on the dev set, we opted to use it for the official runs, leveraging our established models. The ensembles using this straightforward unweighted voting approach, combining embedding-based, prompting, and Agentic Workflow models, outperformed all individual models.

\subsection{LLM Selection}
In our experiments during the model development phase, we observed that, in general, GPT4 performed better on the dev dataset than Claude3 did. Claude3 tended to be too insensitive to non-neutral emotions despite having higher precision on them. Based on this observation, we decided to use GPT4 as the main LLM for our models in the final evaluation on the test dataset while keeping using Claude3 as the second LLM that performed double check on the "Neutral" emotion in the Zero-Shot with Explanation and Correction model.

\section{Results}
In terms of individual models, the Zero-Shot (gpt4o) with Explanation and Correction model (ZSEC-gpt4o) achieved the best performance, achieving an F1-score of 0.5726. Other models, such as ZSEC-gpt4turbo and Multi-Binary-Classifier Agentic Workflow (MBCAWF), also perform competitively, with F1-scores exceeding 0.55. The overall system performance further improved through the use of Agentic Workflow and ensemble methods. Notably, the Ensemble-19 model achieves the highest F1-score of 0.6046 on the test dataset, outperforming the EXALT baseline by approximately 0.17 F1-score, ranking second. Results for all submitted single models, Agentic Workflow models, and ensembles for the emotion detection task are presented in Table \ref{all-results}, with baseline results provided by the EXALT organizers. 

Emotion prediction often differs substantially across emotion labels, and that variation is reflected in our systems as well. Per-emotion F1-scores achieved by our top-performing model, Ensemble-19, illustrate a difference of up to 0.26 F-score, between the highest scoring emotions ( 0.73 for "Anger" and  0.72 for "Neutral") and the lowest scoring emotions - only 0.47 for "Love" and 0.53 for "Fear". This discrepancy underscores the uneven performance of classification across emotion categories and the continuing challenges of this task. The per-emotion F1-scores achieved by Ensemble-19 are shown in Appendix \ref{F}.

\begin{table}[h!]
\centering
\begin{tabular}{llll}
\hline
\textbf{Models} & \textbf{F1-score} & \textbf{Precision} & \textbf{Recall}\\
\hline
EXALT Baseline & 0.43 & 0.43 & 0.44\\    
ZSEC-gpt4turbo & 0.55 & 0.55 & 0.58 \\
ZSEC-gpt4o & 0.57 & 0.56 & 0.60 \\ \hline \hline
MBCAWF & 0.56 & 0.56 & 0.59 \\
MIAWF-3\tablefootnote{built on ZSEC-gpt4o and Ensemble-9} & 0.59 & 0.59 & 0.61\\
MIAWF-5\tablefootnote{built on MIAWF-4 (which is built on MIAWF-3 and Ensemble-8) and Ensemble-8} & 0.60 & 0.59 & 0.62\\ \hline \hline
Ensemble-9\tablefootnote{Ensemble of 9 models (see Table \ref{ensembles} in Appendix \ref{B})} & 0.59 & 0.59 & 0.61\\
Ensemble-8\tablefootnote{Ensemble of 8 models (see Table \ref{ensembles} in Appendix \ref{B})} & 0.60 & 0.60 & 0.62\\
Ensemble-17\tablefootnote{Ensemble of 17 models (see Table \ref{ensembles} in Appendix \ref{B})} & 0.60 & 0.60 & 0.62\\
Ensemble-19\tablefootnote{Ensemble of 19 models (see Table \ref{ensembles} in Appendix \ref{B})} & \textbf{0.60} & \textbf{0.60} & \textbf{0.62}\\
\hline
\end{tabular}
\caption{\label{all-results}
F1-score, precision and recall on the test dataset including the EXALT baseline results.}
\end{table}

\section{Discussion}
There are a few findings that we would like to share. First of all, the effectiveness of explanation and correction over simpler prompting which was found on the dev dataset was replicated on the test dataset. These improvements are detailed in the experiments in Appendix \ref{G}. Secondly, our manual error analysis (Appendix \ref{D}) highlighted 
the subjective nature of the emotion recognition task.  Humans may 
disagree, and the explanation provided by the models may also be reasonable, even in some cases where they do not match the gold standard. 
Thirdly, the justification provided by LLMs could potentially aid the explainability of the outputs. Inspection of automatically generated explanations often showed partial translations, which could be helpful in the cross-lingual setting (Appendix \ref{E}).

Additionally, the tweet data employed in this study are drawn from six high-resource languages. Consequently, it is uncertain whether the models would produce comparable results for lower-resource languages. Further exploration is needed by applying the same methodology to tweets in diverse low-resource languages.

\section{Conclusion}
In this study, we have highlighted the potential of the Agentic Workflow method to enhance emotion detection performance of LLMs on multilingual tweets. Moreover, explicitly prompting LLMs to provide explanations for their decisions not only improves decision-making accuracy but also can aid human comprehension of their decisions. We firmly believe that explainability plays a crucial role in real-world applications by providing insight into the operations of these complex systems. 

At the same time, we should be cautious about the risk associated with using LLMs in subjective tasks, since they may be incorrect but appear confident. Looking ahead, we envision exploring the application of Agentic Workflows across a broader spectrum of fields within sentiment analysis and the wider NLP domain.

\section{Limitations}
The OpenAI and Anthropic models used in this work are closed-source and may continue undergoing reinforcement learning from human feedback (RLHF). Given this situation and the inherent non-deterministic nature of LLMs, reproducing the exact inference results may be challenging. A second issue of using LLMs was that occasionally outputs would be nonsensical, making manual post-processing almost unavoidable. In such case, we simply replaced the problematic outputs, of which the format was not "explanation + emotion label", with "Neutral" labels. Thirdly, due to cost and time constraints, we were unable to perform formal significance tests. Therefore, the results and findings presented in this paper are based on empirical observations from the experiments we conducted. Finally, the model latency of LLMs was quite high for inference on each instance, especially when the raw output contained more text. During the evaluation phase, we broke down the test dataset into multiple parts and parallelized the inference to speed up the process.

% Bibliography entries for the entire Anthology, followed by custom entries
\bibliography{anthology,custom}
% \bibliographystyle{acl_natbib}
% Custom bibliography entries only
%\bibliography{custom}

\appendix
\section{Embedding-based Models}\label{A}
\subsection{K-Nearest Neighbors (KNN)}
We explored KNN in the emotion detection sub-task chosing high-dimensional sentence embeddings as our classification input for our categorical emotion label output. We were motivated by the findings in \citet{yin-shang-2022-efficient} which, although only calculated on English datasets, yielded high-efficiency, high-performing results even when only KNN was used for emotion classification. In our study, we compared KNN performance on OpenAI \& TwHIN-BERT embeddings and found TwHIN-BERT outperformed OpenAI in dev-set F-1 score.
\subsubsection{OpenAI KNN Parameters}
We ran experiments on the dev dataset with different $k$ values (from $1$ to $10$) and with different embedding sizes for both OpenAI embedding models. The setting with best F1 score on the dev dataset was $k=6$ with embedding size 256 using \texttt{text-embedding-3-large} model provided by OpenAI.
\subsubsection{BERT KNN Parameters}
TwHIN-BERT was selected because of its ability to project sentences cross-linguistically onto the same embedding space, fine-tuned on Tweet data. We selected the $k$ (in between 1 and 20) for BERT-KNN using 5-fold cross-validation F1 score on the training data and validation F1 score on the dev set. We were able to identify $k=3$ as offering the best performance on both the training and dev set. 
\subsection{Bidirectional LSTM (BiLSTM)}
Bidirectional Long Short-Term Memory (BiLSTM) has been shown to be capable of capturing the long-range contextual information needed for emotion classification of short messages \citep{bilstmEmotion}. BiLSTM with attention, when applied to a similar implicit emotion classification task for WASSA2018 \cite{zhou-wu-2018-nlp}, yielded competitive performance across emotion classes. Our BiLSTM took as input batches of 280 x 1024 length vectors, where 280 was the BERT tokenizer's padded max sequence length for a sentence and 1024 was the length of each token's TwHIN-BERT embeddings. Then, after feeding the input through a BiLSTM with 256 total hidden cells, we applied an attention layer over all 280 tokens to produce the emotion label for a sentence. The model used an Adam Optimizer coupled with a sparse categorical cross-entropy loss function during training on 90\% of the training data. The remaining 10\% was set aside as validation data and model training stopped after 3 epochs of no validation data loss improvement. The final model's attention layer was then used to produce Numerical Trigger scores for Task 3 as a metric to assess how much individual tokens contributed to a sentence's emotion. Words that had been split into multiple tokens were recombined before outputting these numerical trigger scores.
\section{Information of Different Ensembles}\label{B}
The information are detailed in Table \ref{ensemble_types} and Table \ref{ensembles}.
\begin{table}[h!]
\centering
\begin{tabularx}{0.52\textwidth}{|X|X|X|X|}
\hline
\textbf{Ensemble Types} & \textbf{F1-score} & \textbf{Precision} & \textbf{Recall}\\
\hline
Unweight-ed voting & 0.61 & 0.63 & 0.60\\
\hline
Weighted voting (weighted according to F1-score) & 0.60 & 0.61 & 0.60\\
\hline
Agentic Workflow (GPT4) & 0.49 & 0.51 & 0.52\\
\hline
Agentic Workflow (Claude3) & 0.50 & 0.51 & 0.51\\
\hline
\end{tabularx}
\caption{\label{ensemble_types}
F1-score, precision and recall for all ensemble types on the dev dataset.
}
\end{table}

\begin{table}[h!]
\centering
\begin{tabularx}{0.5\textwidth}{|>{\hsize=0.5\hsize}X|>{\hsize=1.5\hsize}X|}
\hline
\textbf{Ensembles} & \textbf{Base Models}\\
\hline
Ensemble-8 & MIAWF-3, BERT-KNN, ZSEC-gpt4o, FewShot, FineTuneGPT, ZeroShot, OpenAI-KNN, MBCAWF \\
\hline
Ensemble-9 & MIAWF-2, BERT-KNN, ZSEC-gpt4o, FewShot, FineTuneGPT, ZeroShot, OpenAI-KNN, MBCAWF, Explain\_turbo  \\
\hline
Ensemble-17 & 5 MIAWF models (with different source models), 5 ZSEC models (with the same prompts), BERT-KNN, FewShot, FineTuneGPT, ZeroShot, OpenAI-KNN, MBCAWF, BiLSTM  \\
\hline
Ensemble-19 & 5 MIAWF models (with different source models), 5 ZSEC models (with the same prompts), BERT-KNN, FewShot, FineTuneGPT, ZeroShot, OpenAI-KNN, MBCAWF, BiLSTM, Ensemble-8, Ensemble-17  \\
\hline
\end{tabularx}
\caption{\label{ensembles}
Composition of all submitted ensembles on the test dataset.
}
\end{table}

\section{Prompts}\label{C}
\subsection{Fine-tuning GPT3.5}
\textbf{System:} As a supportive assistant specialized in tweet classification, you're tasked with determining the emotion conveyed in a given tweet. Utilizing your intuitive understanding, analyze the sentiment of the provided tweet. Your response should be just one word, choosing one emotion from these 6 emotions: Love, Joy, Anger, Fear, Sadness, Neutral.

\subsection{Zero-Shot and Few-Shot}
\textbf{System:} You are a helpful assistant designed to output classification results.\\
\textbf{User:} Suppose there are six emotions: Love, Joy, Anger, Fear, Sadness, Neutral. Use your instinct, what is the emotion of the following tweet: '\{tweet\_text\}'. Your response must be just one label from the six labels. Please do not output anything else.\\
\textbf{Assistant (only needed for few-shot):} \{label\} 

\subsection{Zero-Shot with Explanation}
\textbf{System:} You are an expert who takes an input tweet and outputs an emotion classification label among the following emotion labels: Love, Joy, Anger, Fear, Sadness, Neutral. Your output should start with the explanation and end with the emotion label. Explanation and emotion label should be separated by ||. Do not output newlines.\\
\textbf{System (only needed for correction):} You are an expert in checking emotion in tweets. There are six emotions 'Love, Joy, Anger, Fear, Sadness, Neutral'. You will be presented with a tweet that others think is '\{emotion\}'. Output '\{emotion\}' if you agree with that; otherwise, output one emotion label from other emotions that describes the emotion of the tweet the best. Your output should start with the explanation and end with the emotion label. Explanation and emotion should be separated by ||. Do not output newlines. \\
\textbf{User:} What is the emotion label of this tweet '\{tweet\}'?

\subsection{Multi-Iteration Agentic Workflow}
\textbf{System:} As an expert specialized in tweet classification, you're presented with a tweet and two emotion labels: "\{emotion1\}" and "\{emotion2\}". Drawing upon your intuitive understanding, assess the emotion of the tweet provided. Your response should be either "\{emotion1\}" or "\{emotion2\}". If the two emotion labels are identical, return either one of them.

\subsection{Multi-Binary-Classifier Agentic Workflow}
\textbf{System (Binary Classifier):} You are an expert in detecting '\{emotion\}' emotion in tweets. You will be presented with a tweet. Output 'yes' if you detect '\{emotion\}' emotion in the tweet; otherwise, output 'no'. Your response should only contain 'yes' or 'no'. No other output is allowed. \\
\textbf{System (Neutral Emotion Check):} You are an expert in checking emotion in tweets. You will be presented with a tweet that others think is 'Neutral'. Output 'Neutral' if you agree with that; otherwise, output one emotion from \{emotions\} that describes the emotion of the tweet the best. In the latter case, your response should only contain one of \{emotions\} and no other output is allowed. \\
\textbf{System (Agentic Workflow):} You are an expert in choosing emotions for tweets. You will be presented with a tweet. Pick one emotion from \{emotions\} that describes the emotion of the tweet the best. Your response should only contain one of {emotions}. No other output is allowed. \\

\section{Empirical Error Analysis for Zero-Shot with Explanation}\label{D}
There are a lot of tweets of which the LLM got the labels incorrect while providing reasonable explanation to justify its own predictions. We are sharing a few of them here from both dev dataset and test dataset. From these samples shared, it is natural for us to think that the LLM predicted the labels of the tweets correctly after reading the explanation. 
\subsection{Samples from Dev Dataset}
\textbf{Tweet 1:} @user His performances at Hull don’t lie , he’s a good player . I remember reading somewhere that he went to Slav multiple times to say he’s being played out of position and was told “you’ll play where I play you”

\textbf{Gold Label for Tweet 1:} Joy

\textbf{Predicted Label for Tweet 1:} Neutral

\textbf{Explanation for Tweet 1:} The tweet reflects a tone of support and mild frustration directed towards the way a player was managed, mentioning how the player's good performances are evident and recalling an instance where the player was dismissed by the manager when raising concerns about their position. The overall sentiment seems to express a level of criticism towards the manager while recognizing the player's talent. There's a hint of negativity due to the mismanagement, yet it's overridden by acknowledgment of the player's ability. 

\textbf{Tweet 2:} @user Fair enough . No 6 - 0 loss is ever acceptable .

\textbf{Gold Label for Tweet 2:} Sadness

\textbf{Predicted Label for Tweet 2:} Anger

\textbf{Explanation for Tweet 2:} The tweet expresses dissatisfaction and a negative sentiment as it highlights that a 6-0 loss is never acceptable. The phrase "Fair enough" might initially seem accommodating, but it's followed by a statement of strong disapproval. This compels us to classify the emotion of the tweet as Anger. 

\textbf{Tweet 3:} yeah my ass is going wherever jasons going lmao

\textbf{Gold Label for Tweet 3:} Neutral

\textbf{Predicted Label for Tweet 3:} Joy

\textbf{Explanation for Tweet 3:} The tweet 'yeah my ass is going wherever jasons going lmao' expresses amusement and lightheartedness, likely in a joking or playful context due to the use of "lmao" (laughing my ass off). This shows joy.

\textbf{Tweet 4:} Finna drop another track

\textbf{Gold Label for Tweet 4:} Neutral

\textbf{Predicted Label for Tweet 4:} Joy

\textbf{Explanation for Tweet 4:} The tweet 'Finna drop another track' suggests excitement and anticipation about sharing new music. The conveyed emotion is positive, reflecting happiness or pride associated with creation or expression in music. Therefore, the appropriate emotion label is Joy.

\textbf{Tweet 5:} @user Just from reading the article it would seem sessions and Page will Be indicted soon if the Ruskies don’t get them first

\textbf{Gold Label for Tweet 5:} Sadness

\textbf{Predicted Label for Tweet 5:} Fear

\textbf{Explanation for Tweet 5:} The tweet suggests a negative anticipation, possibly with a hint of fear or concern, regarding potential indictments and external threats. The mention of "indicted soon" and "Ruskies don’t get them first" conveys a worrying and slightly fearful tone, with an element of political tension. Therefore, the most fitting emotion classification is Fear. 
\subsection{Samples from Test Dataset}
\textbf{Tweet 1:} @user Yea he found it hilarious afterwards 

\textbf{Gold Label for Tweet 1:} Neutral

\textbf{Predicted Label for Tweet 1:} Joy

\textbf{Explanation for Tweet 1:} The tweet describes someone finding something hilarious, which indicates happiness and amusement.

\textbf{Tweet 2:} Aplem is doing a \$ 50 and \$ 20 Visa Card giveaway ! Check it out and dont miss your chance to get one ! http

\textbf{Gold Label for Tweet 2:} Neutral

\textbf{Predicted Label for Tweet 2:} Joy

\textbf{Explanation for Tweet 2:} The tweet expresses excitement and anticipation about a giveaway, which are indicators of a positive and joyous emotion.

\textbf{Tweet 3:} @user Me too . Namely women .

\textbf{Gold Label for Tweet 3:} Joy

\textbf{Predicted Label for Tweet 3:} Neutral

\textbf{Explanation for Tweet 3:} The tweet seems neutral and lacks emotional context. It is a mere statement without any indication of strong feelings. 

\textbf{Tweet 4:} Debate over whether that should have been a penalty , but the bottom line is Fernanmendi bundled into Agutip - you cannot do that .

\textbf{Gold Label for Tweet 4:} Anger

\textbf{Predicted Label for Tweet 4:} Neutral

\textbf{Explanation for Tweet 4:} The tweet discusses a controversial situation in a sports event and the rules associated with it but does not express any overtly strong emotional response.

\textbf{Tweet 5:} back at it again http

\textbf{Gold Label for Tweet 5:} Joy

\textbf{Predicted Label for Tweet 5:} Neutral

\textbf{Explanation for Tweet 5:} The tweet "back at it again http" doesn't provide enough context or explicit emotional content. It appears to be a neutral statement about returning to an activity.

\section{Zero-Shot with
Explanation on Non-English Data}\label{E}
The following samples include tweets in different languages from both the dev dataset and test dataset and our model predicted the emotions labels for these tweets correctly. With the explanation, we were somewhat able to see why different labels were assigned without knowing the languages of the tweets. Note that we identified the languages of the sample tweets using Google Translate for the samples presented below.
\subsection{Samples from Dev Dataset}
\textbf{Tweet 1 (Spanish):} Los tipos siguen sin traer la suya al país ! ¿Explícame como esperan que esto se calme si todos sabemos que ustedes no confían en ustedes ?

\textbf{Gold Label for Tweet 1:} Anger

\textbf{Predicted Label for Tweet 1:} Anger

\textbf{Explanation for Tweet 1:} The tweet expresses frustration and dissatisfaction with some people not bringing something important to the country and questioning their reliability, indicating a sense of distrust and annoyance. This feeling aligns with feelings of anger.

\textbf{Tweet 2 (French):} Après faut pas trop leur en demander , c ’ est pas les plus futés qu ’ il y a là dedans http

\textbf{Gold Label for Tweet 2:} Anger

\textbf{Predicted Label for Tweet 2:} Anger

\textbf{Explanation for Tweet 2:} The tweet "Après faut pas trop leur en demander, c'est pas les plus futés qu'il y a là dedans" implies a dismissal or a derogatory comment about the intelligence of others, suggesting a lack of respect or slight frustration. This conveys a sense of disdain or irritation towards the subjects being discussed, which aligns with the emotion of anger.

\textbf{Tweet 3 (Dutch):} Weersverwachting : Vandaag wordt het in Overlangel Zwaar bewolkt , lichte regen , min . 4 graden en niet warmer dan 1 . . .

\textbf{Gold Label for Tweet 3:} Neutral

\textbf{Predicted Label for Tweet 3:} Neutral

\textbf{Explanation for Tweet 3:} The tweet is a simple weather forecast stating the conditions as heavily cloudy with light rain and mentions temperatures. It has a factual tone without any personal feelings or emotions. Therefore, the emotion label for this tweet is Neutral.

\textbf{Tweet 4 (Russian):} 
\selectlanguage{russian}
Уважаемые коллеги , пришла ко мне моя знакомая , случай которой я описывала . . . http http

\selectlanguage{english}
\textbf{Gold Label for Tweet 4:} Neutral

\textbf{Predicted Label for Tweet 4:} Neutral

\textbf{Explanation for Tweet 4:} The tweet provides minimal context or emotional indicators, primarily functioning as a statement of fact about a colleague visiting and mentioning a previous situation. The contents lack any clear emotional language or sentiment. Therefore, it is classified as neutral. 
\subsection{Samples from Test Dataset}
\textbf{Tweet 1 (Spanish):} Te quiero pa ’ mi

\textbf{Gold Label for Tweet 1:} Love

\textbf{Predicted Label for Tweet 1:} Love

\textbf{Explanation for Tweet 1:} The phrase "Te quiero pa' mi" translates to "I want you for myself" in English and expresses a desire to be with someone, often imbued with affectionate and loving intent. 

\textbf{Tweet 2 (French):} un boom dans le monde du jeu vidéo ? http

\textbf{Gold Label for Tweet 2:} Neutral

\textbf{Predicted Label for Tweet 2:} Neutral

\textbf{Explanation for Tweet 2:} The tweet asks a question about a significant event in the world of video gaming and includes a link, suggesting curiosity or interest but not clearly expressing any strong emotion. 

\textbf{Tweet 3 (Dutch):} Zelfkennisbieb is uit haar functie ontheven . Drie dagen ontspanning . Daarna weer \# inzicht opdoen . Kom jij dan lenen ? http

\textbf{Gold Label for Tweet 3:} Neutral

\textbf{Predicted Label for Tweet 3:} Neutral

\textbf{Explanation for Tweet 3:} The tweet mentions self-awareness being relieved from duty, followed by three days of relaxation and then gaining insight again. The overall tone is neutral and informative without strong emotional expressions.

\textbf{Tweet 4 (Russian):} 
\selectlanguage{russian}
14 . Выгляжу лучше , чем когда - либо , довольна собой .

\selectlanguage{english}
\textbf{Gold Label for Tweet 4:} Joy

\textbf{Predicted Label for Tweet 4:} Joy

\textbf{Explanation for Tweet 4:} The tweet translates to "14. I look better than ever, pleased with myself." which expresses self-satisfaction and happiness. 

\section{Per-emotion F1-scores Achieved by Ensemble-19}\label{F}
\begin{table}[h!]
\centering
\begin{tabularx}{0.49\textwidth}{|X|X|X|X|X|}
\hline
\textbf{Emotion Labels} & \textbf{F1-score} & \textbf{Precis-ion} & \textbf{Recall} & \textbf{Support}\\
\hline
Love & 0.47 & 0.55 & 0.41 & 190\\
\hline
Joy & 0.63 & 0.55 & 0.74 & 433\\
\hline
Anger & 0.73 & 0.76 & 0.70 & 614\\
\hline
Fear & 0.53 & 0.44 & 0.68 & 77\\
\hline
Sadness & 0.55 & 0.56 & 0.54 & 270\\
\hline
Neutral  & 0.72 & 0.76 & 0.68 & 916\\
\hline
\end{tabularx}
\caption{
Per-emotion F1-scores of the Ensemble-19 model.
}
\end{table}

\section{Effectiveness of Explanation and Correction}\label{G}
Note that during the evaluation phase, we only ran ZeroShot (gpt-4-turbo), ZSE (gpt-4o) and ZSEC (gpt-4o), of which the results are shown in Table \ref{explanation-correction-table}. For ZeroShot (gpt-4o), we ran it after the gold labels were released for the test dataset. We noticed that there was a nonnegligible improvement on the evaluation metrics with ZeroShot (gpt-4o) comparing to ZeroShot (gpt-4-turbo). It remains uncertain to us whether the gpt-4o model has been updated since the evaluation phase, because due to cost and time constraints, we were unable to re-run the ZSE (gpt-4o) and ZSEC (gpt-4o) models again after the gold labels were released for the test dataset.
\begin{table}[h!]
\centering
\begin{tabularx}{0.53\textwidth}{|X|X|X|X|X|}
\hline
\textbf{Models} & \textbf{F1-score} & \textbf{Precis-ion} & \textbf{Recall} & \textbf{Accur-acy}\\
\hline
ZeroShot (gpt-4-turbo) & 0.5459 & 0.5539 & 0.5682 & 0.6028\\
\hline
ZeroShot (gpt-4o) & \textbf{0.5732} & \textbf{0.5685} & 0.5813 & 0.6164 \\
\hline
ZSE (gpt-4o) & 0.5723 & 0.5664 & 0.589 & 0.6232 \\
\hline
ZSEC (gpt-4o) & 0.5726 & 0.5631 & \textbf{0.5953} & \textbf{0.624} \\
\hline
\end{tabularx}
\caption{\label{explanation-correction-table}
Precision, Recall, F1-scores and Accuracy on the test dataset for ZeroShot, ZSE and ZSEC (correction on "Neutral").
}
\end{table}

\end{document}